# Artificial Intelligence for Health Message Generation: Theory, Method, and an Empirical Study Using Prompt Engineering

Sue Lim[1], & Ralf Schmälzle[1]

[1]*Department of Communication, Michigan State University, East Lansing, USA*

## Abstract

This study introduces and examines the potential of an AI system to generate health awareness messages. The topic of folic acid, a vitamin that is critical during pregnancy, served as a test case. Using prompt engineering, we generated messages that could be used to raise awareness and compared them to retweeted human-generated messages via computational and human evaluation methods. The system was easy to use and prolific, and computational analyses revealed that the AI-generated messages were on par with human-generated ones in terms of sentiment, reading ease, and semantic content. Also, the human evaluation study showed that AI-generated messages ranked higher in message quality and clarity. We discuss the theoretical, practical, and ethical implications of these results.

Consider the following two messages: *"Every woman needs #folicacid every day, but even more women need it in their first trimester! Folic acid is an essential nutrient needed for the formation of new blood cells in the womb, as well as the developing nervous system"*, and *"The risk of neural tube defects is reduced if women consume 400 micrograms of #folate every day before and during early pregnancy."* Could you tell which of these messages was written by an AI system? We will resolve this question at the end of this paper, which will discuss the theoretical potential, practical, and ethical implications of AI for communication science, here focusing first on the generation of health awareness messages[1].

The paper is structured as follows: First, we introduce the theoretical underpinnings of AI-based message generation. The following section will discuss the application context, health communication, and specifically health awareness campaigns on social network sites. The third section will bring these two research streams together, discussing how AI language models can generate health awareness messages and introducing recent research in this area. We will then present the study with its hypotheses, methods, and results. Finally, we will discuss the results and their implications.

---

[1] Importantly, we anticipate that the scope and impact of AI in communication will go far beyond health awareness messaging, including AI-based health persuasion (focusing on attitude and behavior change), AI in political communication, advertising, as well as interpersonal domains. However, this paper focuses exclusively on the generation of brief messages intended to raise awareness; we view this as a high-impact and decidedly pro-social domain in which communication and AI can connect. Throughout the paper, we will emphasize that at this stage, influencing attitudes and behavior via AI-generated messaging systems is a possibility (and likely already ongoing), but this empirical research paper focuses only on the proximal goal of raising awareness.

## AI Message Generation: Theory, Method, and Relevance for Communication

Theoretical breakthroughs in deep learning (DL), most of which happened between 2010 and 2020, have equipped computers with impressive capacities. In brief, deep learning uses artificial neural networks that are stacked across multiple layers (hence: deep) and are trained to perform specific tasks (hence: learning), such as automatic speech and image recognition (e.g., for voice dictation, labeling objects in photographs, etc.). As of now, DL-fueled applications are integrated in every cell phone and computer (Chollet, 2021; Hassabis et al., 2017; Schmidhuber, 2015). While these capacities were initially geared more towards perceptual and receptive tasks, like speech and image recognition, the field soon expanded to include generative, productive tasks, like text or image synthesis. For example, every text processing software now includes the ability to auto-complete words from their beginnings, or even suggest entire sentence continuations.

To develop a brief, non-technical intuition of the underlying principles, consider the sentence "I take my coffee with cream and …". Most humans find it easy to predict the next word as "sugar". Logically, given the semantic context set up by the preceding words, the word sugar is very likely - certainly far more likely than e.g., "socks". The way in which so-called language models (artificial neural networks for dealing with natural language) are trained is actually quite similar to this example. In particular, researchers use large amounts of text from the internet, automatically mask out individual words, feed the neural network the remaining words, and train it to correctly predict the masked word. Once this training process is complete, the model will have learned the hidden statistical relationships between words in a large corpus of natural language. Then, if prompted with a sentence, like "Thou shalt …", it will be able to predict that the likely next word is "not". Of note, this principle can be applied over multiple scales, from letters in words (like when a word starts with "th" it is likely that the next letter will be an "e") to words in sentences (like in the example above), and beyond.

A subfield of DL called generative deep learning (generative DL; Chollet, 2021; Goodfellow et al., 2016), takes these ideas to the next level by using pre-trained language models for text generation (Gatt & Krahmer, 2018; Keskar et al., 2019; Tunstall et al., 2022). Specifically, one can take a starting sequence of words, feed this as a prompt into the model, and then use the most likely next word to continue the sentence. This is exactly what your cell phone or email programs do. Critically, however, one can continue this process by feeding the updated sequence into the LM (i.e., the starting sequence along with the generated word), generating a likely next word again, and so forth. This way, we can generate text that appears natural to humans. To summarize, deep learning technologies have matured up to a degree where they can generate complex textual content (Rashkin et al., 2020; Tunstall et al., 2022) and readers have certainly encountered impressive examples of such text generation systems. Although many limitations exist (Bender & Koller, 2020), which we will discuss below, these advances are very relevant for communication science, and they have many theoretical implications.

Put simply, if these systems can generate messages[2], then communication scientists have a theoretical obligation to study them, including their principles, effects on humans, limitations, and so forth. To unpack this, we refer to three interconnected reasons. First, the act of message generation is generally understudied and undertheorized within communication science: Coming up with a novel message is to a non-negligible degree still a creative endeavor (Gelernter, 2010) that is accomplished via intuition (Hodgkinson et al., 2008). Although there are certain scientific principles and some theories of message design (Cho, 2011; Greene, 2013; Witte et al., 2001), we must acknowledge that there is a gap in our understanding of how we come up with new messages - or else there would already be the kinds of message-generating machines that we propose below. Evidence for this can also be seen in the fact that message generation is by far not as widely studied as message analysis: there are literally thousands of studies examining responses to and effects of health messages, but message creation is rarely studied, at least not in the way we propose to study here[3]. Even when content analyses examine the ingredients of messages, researchers only rely on existing messages but again do not study how they are generated.

Second, theory and method evolve in a synergistic relationship (Greenwald, 2012) in which technological advances promote new theory and theoretical advances provide the basis for new methods. Applying this to the current context, we see that the new abilities to generate messages (see the paragraphs above) expand our theoretical understanding of what has so far remained enigmatic, namely how new, coherent messages can be created. In this sense, putting method and theory in opposition is - although common - actually misleading.[4]

Third, as methods promote theory and vice versa, we often see the emergence of entirely new fields of research. For instance, when computers came into contact with communication, the field of CMC arose, which in turn stipulated theory. Similarly, theory and methods from communication and neuroscience are converging, giving rise to the field of communication neuroscience (Huskey et al., 2020; Schmälzle, 2022; Schmälzle & Meshi, 2020). Together, these three arguments underscore why the nexus of AI and communication is a theoretical wellspring rather than "a method in search of a problem":[3] DL-approaches for text generation help explain a henceforth mystic process and they provide us with a principled approach to generate thousands of novel messages, which we can analyze. This development is likely going to impact many fields of communication, including health communication, to which we will turn next.

---

[2] We acknowledge that definitions of communication, message, and theory can vary, and there are many longstanding debates about what counts and what does not count as communication, message, or theory. Clearly, the field of NLP is closer to engineering and computer science and has to a large degree ignored communication science (e.g., Bender & Koller, 2021). Nevertheless, there have always been theoretical connections, like between computer-mediated communication and voice assistants, or using computational methods to analyze communication. The advent of text generation methods - although they still certainly lack many if not the most central characteristics of human communication (e.g., intent, goals, pragmatics, etc.) - is clearly bringing the fields into closer contact. As such, we believe that it will be in communication scientists' best interest to engage with this development, which is already influencing key aspects of communication in the 21st century.

[3] To be clear, one could certainly view the entire literature on rhetoric as being focused on message production and delivery, like the notion of 'inventio' in Aristotle's theory, which can be considered the birthplace of communication science. Likewise, work on communication goals and strategic communication also have action- or production-oriented elements. However, these approaches are geared towards more abstract levels, and they are all silent about the actual nuts and bolts of message generation.

[4] Paraphrasing Lewin, who once said that "There's nothing so practical as a good theory" (1943), Greenwald (2012) states that "There is nothing so theoretical as a good method." History clearly supports both statements, like when the development of telescopes massively propelled astronomical theories, or the cross-pollination of neuroscience and mathematical theories in the 1940s, which paved the way for the deep learning revolution in the last decade.

**Health Communication, Awareness Campaigns, and the Message Generation Bottleneck**

The need to communicate about health is clear, compelling, and constantly high: About 60% of worldwide deaths and a large share of the global burden of disease are due to preventable factors (Ahmad & Anderson, 2021; Giles, 2011; Mokdad et al., 2004). By using health communication to inform and influence individuals, we hope to be able to reduce these numbers and help people to live healthier and happier lives. The COVID-19 pandemic has made this all particularly salient, although there are many other topics beyond infectious disease and vaccinations where health communication comes into play (e.g., general health promotion and disease prevention, nutrition and lifestyle, substance use, risk behaviors, etc.; Thompson & Harrington, 2021).

Health communication research is complex and multifaceted, so this section will only focus on the topic of health-related mass media campaigns (Atkin & Silk, 2008.; Rice & Atkin, 2012). Even within this more confined area, we will only focus on awareness messaging. Health awareness campaigns are aimed at increasing the public's knowledge about a particular health issue and ways to prevent it. Though much of health communication research examines attitude and behavioral change, increasing the public's knowledge about the health issue sets the foundation of any health campaign[5]. One must know about the health problem to have an attitude or act upon it. Existing literature provides examples of awareness influencing attitudes and behavior (e.g., Yang & Mackert, 2021). In addition, simply increasing public knowledge about a health issue can help alleviate the extent of the problem. Thus, campaigns' focus can also be to engage the public and increase knowledge (Bettinghaus, 1986).

Mass communication channels, particularly social media platforms, provide cost-effective tools to spread awareness about health issues to large audiences (Shi et al., 2018; Willoughby & Noar, 2022). On social media, people can connect with others and have social networks that extend beyond geographical boundaries. Users can share information and pass it on to another through their social networks. Social networks, then, allow information to be disseminated to a large number of people at a rapid pace. Existing literature has shown that social media can be effectively used to raise awareness on health prevention issues such as Covid (Chan et al., 2020), skin cancer (Gough et al., 2017), or the famous ALS ice-water bucket challenge (Shi et al., 2018; Wicks, 2014). In addition, the true power of social media is when content becomes "viral", achieving a significantly high amount of awareness (e.g., ALS ice-water bucket challenge). Health researchers have already begun examining how to leverage this phenomenon on social media to expand the research of health messages (Kim, 2015; Thackeray et al., 2008; Wang et al., 2019). Thus, if leveraged well, communicators can potentially reach billions of people from all over the world through awareness campaigns.

---

[5] Please note that the goal of this study focuses only on generating messages that are capable of raising awareness about health issues. Clearly, additional goals of health communication and persuasion, particularly attitude and potentially even behavior change, are within the realm of possibilities, but these are not the goal of this study. To our knowledge, these goals have not yet been addressed by research and we deliberately chose to focus on the arguably simpler goal of raising awareness and dealing with the complexities of attitudes and behavior at a later point. That said, however, awareness about a health issue is the most obvious and important starting point because people cannot prevent risks of whose existence they are not aware (Bettinghaus, 1986; McGuire et al., 2001; Schmälzle et al., 2017).

On the other hand, the rapid advancements in social media have created additional challenges for health professionals (Shi et al., 2018; Willoughby & Noar, 2022). For one, each social media platform has different characteristics of predominant users, making it a challenge to truly understand what types of people the campaigns will reach on the platforms. Second, features that make content viral on social media platforms constantly change. This dynamic nature of social media presents a significant challenge for health researchers. While large companies may have the financial resources to hire prominent influencers, health campaigners often lack these resources. In addition, crafting health messages often takes time and manpower to conduct formative research, create messages, and then pretest the messages (Rice & Atkin, 2012; Snyder, 2007). By the time messages are created, the content may no longer be interesting enough to the audience. Thus, generating and disseminating quality health content fast enough to match the speed of developments in social media has become a challenging task for campaigners. A message generation system that can generate awareness messages within a few minutes, then, can help reduce the bottleneck of message creation.

## Promoting Folic Acid Awareness as a Test Case for AI Health Message Generation

This study will use folic acid (FA) as a test case to examine whether a state-of-the-art AI system is capable of generating health messages to promote awareness about this important topic. Folic acid, or folate, is a type of vitamin B9 that is essential for new cell generation and building DNA (CDC, 2022; *Folate (Folic Acid)*, 2012; Geisel, 2003). While it is suggested that folic acid may help prevent health issues in adults (e.g., stroke; Wang et al., 2007), it is most known to prevent birth defects such as neural tube defects (NTDs) in newborn babies (CDC, 2022.; Scholl & Johnson, 2000). NTDs refer to defects in the spinal cord, brain, and areas around them, and they are part of the five most serious birth defects in the world (Githuku et al., 2014). This defect is especially dangerous because it could form in the fetus during the early stages of pregnancy, sometimes even before the mother is aware of the pregnancy. To prevent NTDs, CDC recommends that women consume at least 400 mcg of folic acid a day, especially during the early stages of pregnancy (CDC, 2022; Gomes et al., 2016). Despite the significance of consuming folic acid, there is a general lack of awareness regarding this issue. Spreading awareness about the issue can therefore help decrease the rate of birth defects (Green-Raleigh et al., 2006; Medawar et al., 2019).

Extant folic acid awareness campaigns showed some effectiveness in increasing knowledge about folic acid. For example, Rofail et al.'s (2012) review showed that existing folic acid campaigns increased knowledge of sources of folic acid. Other extant literature showed that folic acid awareness campaigns not only increased knowledge but also promoted an increase in folic acid consumption (Amitai et al., 2004). Taken together, folic acid remains an important health topic that suffers from a chronic lack of awareness because too few women of childbearing age are aware of the link between folic acid and NTDs during early pregnancy (Medawar et al., 2019). Thus, folic acid is an ideal health issue to test the value of an AI-message engine.

## Using AI to Generate Health Awareness Messages

The previous section stated the need for health communication in general and social media-based awareness campaigns in particular. In this section, we will shift back to AI-based message generation and review how newly available text-generation technologies have already been applied to health communication in very few prior studies. To our knowledge, only one paper in communication has used NLP models to generate health awareness messages. Schmälzle and Wilcox (2022) introduced a message machine capable of creating new awareness messages after fine-tuning. Fine-tuning refers to retraining a language model with a dataset so the model can generate messages with similar structure and content as the dataset (one can think of it as expert training in a specific domain). Specifically, the authors fine-tuned a GPT-2 based model on messages about folic acid and then used it to generate novel messages about this topic. They compared 30 AI-generated messages against 30 human messages and found that AI-generated messages were on par, and even minimally higher compared to the human-generated messages in terms of quality and clarity.

Though not focused on awareness messaging, one other academic paper has used NLP to generate persuasive messages. Karinshak et al. (2022) used GPT-3, a successor model of GPT-2, to generate Covid-19 pro-vaccination messages. GPT3 enables so-called zero-shot learning or generating messages from prompts without prior fine-tuning. The results showed that participants rated AI-generated messages as more effective, having better arguments, and they affected post-exposure attitudes more strongly. Both examples above show that AI message generation is feasible and promising for health communication.

## The Current Study and Hypotheses

The current study tested people's perception of AI-generated vs. human-generated messages in terms of quality and clarity. Compared to prior work, this study implemented several key innovations: a newer and more powerful language model (Bloom), a novel message generation strategy (prompting), and a more challenging comparison standard (comparing AI-generated messages to the most shared messages).

This study used the Bloom model, the most recent and largest open-source NLP model available to researchers (Bigscience, 2022). In addition to being openly available, Bloom allows for prompting, which refers to inputting the beginning part of a text (e.g., *"A good night's sleep is important because ..."*). Then the machine provides the text output that begins with the prompt (e.g., *"A good night's sleep is important because your body needs to recover"*). Simply put, the prompting technology provides a context, thereby constraining the topic of the generated text to the situation provided in the prompt. Studying the mechanisms and effects of prompting is currently an active topic, known as prompt engineering, which holds much potential for communication scientists (Lin & Riedl, 2021; Liu et al., 2021).

This study also improved the process for selecting human-generated messages as comparison standards for AI-generated ones. Schmälzle and Wilcox's (2022) study randomly selected human-generated messages from the scrapped tweets. In this study, we randomly selected human-generated tweets as well, but this time, from the retweeted messages only. Retweeting represents an objective outcome that is easy to assess and one of the gold standards of messaging success (DellaVigna & Gentzkow, 2010; Rhodes & Ewoldsen, 2013). In other words, retweeted messages contain elements deemed more worthy of sharing by audiences (Dmochowski et al., 2014; Pei et al., 2019). Thus, by comparing AI-generated messages to retweeted human-generated messages, this study examined the feasibility of using the message engine to not simply generate new FA messages, but to generate those that are at least on par with the content people share with others.

With these innovations and modifications, the current study examined the potential of AI language models for health awareness generation. Specifically, we asked two research questions and postulated one hypothesis. The first research question concerned the feasibility of awareness message generation in general and particularly the novel strategy, i.e., using the Bloom foundation model with a prompting approach (RQ1). To address this question, we chronicled the system's requirements, ease of use, and described the setup, speed, computational demand, and general behavior. The second research question asked what general characteristics the generated messages exhibited (RQ2). We addressed this question through a mix of qualitative assessment and detailed computational analyses of the generated messages. Lastly, based on the prior results, we hypothesized that AI-generated messages would be at least on par with the shared human-generated content (H1). To this end, we conducted an online study in which we exposed participants to human- and AI-generated messages and asked them to evaluate the messages in terms of quality and clarity.

## Method

In this section, we will first describe the architecture of the message generation system (called message engine), how the system was used to generate messages (via so-called prompting), and how we evaluated the generated messages via computational and human evaluation methods.

**Description of Bloom, State-of-the-Art NLP Model**

As mentioned above, we used Bloom, the latest and largest open-source multilingual language model. Bloom is powered by a transformer-based ANN architecture that is similar to that of OpenAI's GPT-3. For feasibility, we used the second largest version (7B1) with 7 billion neural network parameters for this study. Bloom 7B1 was trained on 1.5 TB of pre-processed text from 45 natural and 12 programming languages (Bigscience, 2022) including, for example, the Wikipedia and the Semantic Scholar Open Research Corpus (Lo et al., 2019). As previously mentioned, instead of fine-tuning this model as done in previous research, here we used a prompting-based generation strategy. See Figure 1 for a conceptual diagram of message generation via Bloom.

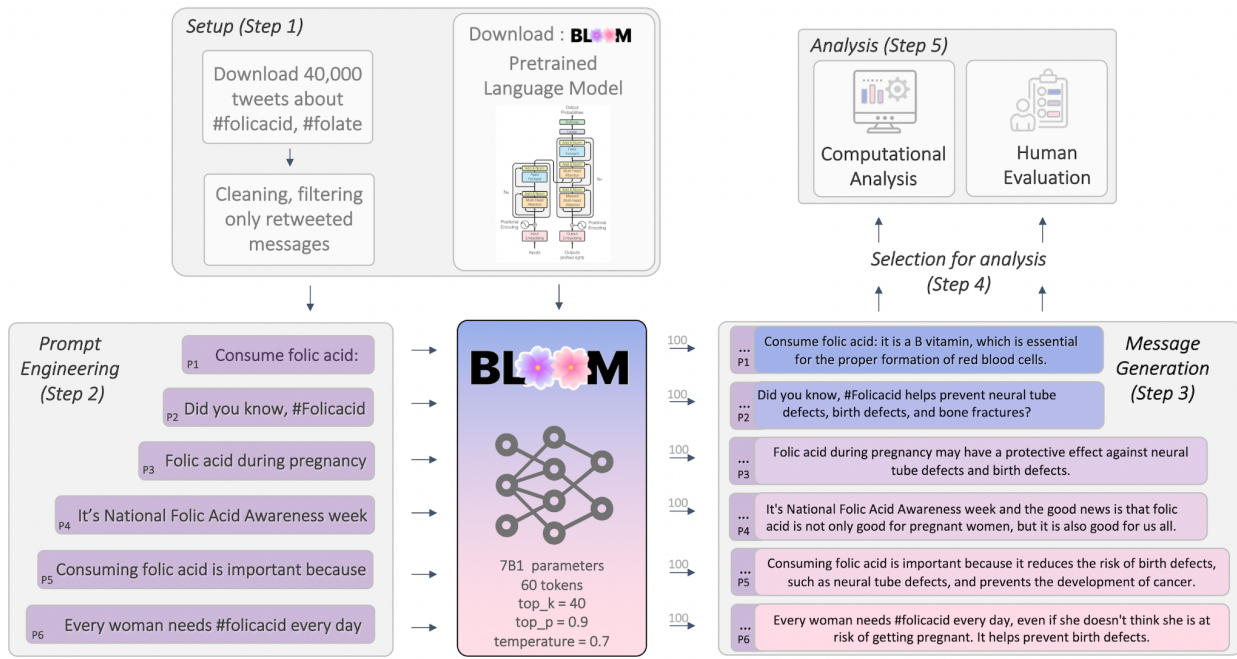

**Figure 1:** Message generation system leveraging Bloom, a large pretrained language model in combination with prompt engineering to generate health awareness messages about folic acid (Vaswani et al., 2017).

**Message Generation Protocol**

*Step 1: Collecting Retweeted Human-Generated Tweets.* The first step consisted of scraping FA messages from Twitter. Using the snscrape package (*Snscrape*, 2021) in python, we collected all tweets with #folic acid or #folate on Twitter (no date constraints). This yielded a base corpus of 42,646 raw messages, which were exported into a csv-file and sorted by retweeted count. The top 50 tweets were selected and filtered for duplicates, non-English text, and promotional content (words buy, order, sale) resulting in a total of 28 tweets. The next two most retweeted tweets were added to the list to obtain the 30 most retweeted human-generated tweets about FA.

*Step 2: Prompt Engineering.* As prompting can significantly influence the generated messages (Liu et al., 2021), five of the six prompts were crafted based on the sentence structure of the most retweeted messages. Starting from the most retweeted message, the beginning phrases of each of the tweets were examined and selected as long as the phrase did not contain too much information about folic acid. Using the transformer python package, we entered the selected phrase as the prompt of the bloom model and examined 30 generated tweets.

The prompts were discarded if we observed the following red flags: too many of the same messages without much informative content, too many names of official organizations or countries, or clearly false information. Some phrases were modified and tested as prompts as well. For instance, the most retweeted tweet was "This is fantastic news, very pleased that the government has finally accepted the strong evidence for this policy. #Folicacid to be added to UK flour in effort to reduce birth defects". Adopting this, we tested the prompt "This is fantastic news! #Folicacid". However, this prompt was discarded because the prompt generated too many messages with official country and organization names. The process ended when five prompts were selected.

Two of the three prompts then were modified by turning the hashtag #folicacid into the words folic acid. This modification was made because we wanted to instruct the model to generate messages about folic acid without restricting the message to tweet format. However, we still kept the structure of the tweet. For example, instead of using "Consume #folicacid:", we modified the prompt to "Consume folic acid:". Then, one more prompt was created based on the general structure of a simple awareness message: "Consuming folic acid is important because". In sum, we used the following six prompts: "It's National Folic Acid Awareness week", "Every woman needs #folicacid every day," "Did you know, #Folicacid", "Consuming folic acid is important because", "Consuming folic acid:" and "Folic acid during pregnancy".

*Step 3: AI Message Generation.* In addition to the varying prompts, the generation command used fixed values for the parameters *maximum result length, sampling, temperature, top-k,* and *top-p*. The *maximum result length* specifies the amount of text to be generated and was set to 60 tokens (approximately 45 words) to roughly match the character limit of Twitter messages. The other parameters allow the precision and randomness of the generated messages. By setting the parameter *do_sample=True*, we instructed the model to use the sampling approach, which means that the model will calculate the conditional probability of each word that was stored via training and randomly selected based on the probability. Sampling was chosen over other methods in order to add some variation into the text generation, modeling after the variations that exist in natural human language (Holtzman et al., 2019; Wolf et al., 2020). We then determined the *sampling temperature*, or how much randomness is allowed in generating the text after the prompt. The *temperature* was set to 0.7 to follow the recommended levels for multiple or longer text generation (Misri, 2021). Next, top-k and top-p sampling strategies were used in combination in order to allow for variation in generated text while preventing words with low calculated probabilities from being selected (*top_k*=40, *top_p*=0.9; Fan et al., 2018; von Platen, 2020). Using these parameters, a total of 600 messages were generated, 100 messages for each of the six prompts.

*Step 4: Selecting AI-generated Messages for Evaluation.* After generating the messages, the following procedure was used to select 30 messages for comparison against human generated messages. Using a random number generator, 10 messages were randomly selected from each of the 100 generated messages. Among the 10 messages, messages were discarded if we found the following exclusion criteria: 1) clear false information, 2) specific references to non-US specific organizations (e.g., NHS, UK, Scotland, Netherlands), 3) references to sources that could not be verified (e.g., "...the recommended intake of 400 Œºg/day is based on a meta-analysis of randomized controlled trials (RCTs)), and 4) recipes that include folic acid. In addition, two of the 60 randomly selected messages were discarded because they included characters in different languages or had repetitive phrases. If more than 5 messages passed the exclusion criteria, then we selected the messages that included more content than hashtags or differed from other included messages. Finally, the selected messages were cleaned (e.g., & changed to &, end part of the messages deleted if the sentence was not complete).

**Methods for Computational Evaluation Study**

The goal of the computational analyses was to examine the similarities in the AI- and human-generated messages. Specifically, we examined the AI-generated and human-generated messages via a number of common text analytic methods in python in R, including N-gram analysis, semantic analysis, readability analysis, topic modeling, and sentiment analysis. Through n-gram analysis and semantic analysis, we hoped to examine the similarities in the word distributions and the general attributes of the sentences. The readability analysis examined how easy the AI-generated messages were to understand compared to human messages (DeWilde, 2020; Flesch, 1946). Using topic modeling (Blei, 2012) and sentiment analysis (Hirschberg & Manning, 2015; Hutto & Gilbert, 2014), we examined the similarities and differences in the discussed topics and the general sentiment of the messages. These analyses were carried out using python and R packages including spacy, textacy, vader, and the sentence-transformers (DeWilde, 2020; Hornik & Grün, 2011; Hutto & Gilbert, 2014; *Industrial-Level Natural Language Processing*, 2022.; Reimers & Gurevych, 2019.).

**Methods for Human Evaluation Study**

*Participants.* Participants were recruited from a study pool and received course credits as compensation for the study, which was approved by the local review board. Responses from participants who completed the survey at an unrealistically fast speed (<4 minutes) or failed to complete the survey were discarded. The final dataset included $N = 120$ respondents, with 70% being female ($n = 84$). Since folic acid is especially significant for stages during pregnancy, the high proportion of female participants fit the purpose of our study. We also conducted a power analysis using the pwr package in R (Champely, 2020) for a one-sample and one-sided *t*-test, with effect size $d = .3$ and significance level $α = .05$. Sample size of 100 was enough to detect significance at the power level of .9. Moreover, we relied on evidence suggesting that a sample of this size is more than sufficient to select best-performing messages as candidates for campaigns (Kim & Cappella, 2019).

*Procedure.* The study was conducted online via Qualtrics, and participants were not told in advance which of the messages were AI-generated or human generated; they were only told that the purpose of the study was to evaluate health messages related to folic acid. Once participants consented to the study, they were asked to evaluate the quality and clarity of the messages in blocks. Message order was randomized within each block and approximately half of the sample started with the first question, the other half started with the second block. The two survey items were adopted from Schmälzle and Wilcox (2022). The quality of the message measure asked, *"How much do you agree that the content and the quality of this message is appropriate to increase public knowledge about folic acid,"* and responses varied from 1 (strongly disagree) to 5 (strongly agree). The clarity of the message measure stated, *"Please evaluate the following messages in terms of whether they are clear and easy to understand,"* with 1 meaning very unclear and 5 meaning very clear.

Survey data were downloaded and further processed in python. Specifically, we computed by-message averages for each question, and also averaged across the blocks of AI- and human-generated messages, respectively. Hypothesis testing was conducted using the scipy-package (with an *α*- level of 0.05) to test for differences between AI- and human-generated messages in each measure. Furthermore, the participant sample was also split by sex (due to the relevancy of folic acid during pregnancy).

# Results

This study tested how feasible and easy it is to use the Bloom message engine to generate clear, effective, and novel awareness messages. In this section, we first discuss the ease of harnessing the technical aspects of the message engine (RQ1). Next, we describe characteristics of the generated messages and present the results from the computational methods (RQ2). Finally, we show how the AI-generated messages compared to retweeted human-generated messages in terms of clarity and quality from the online survey study (H1).

**Feasibility of the System**

The Bloom message engine was much more efficient and simpler to use compared to the GPT-2 model used in the previous study. The main appeal of Bloom is its ability of zero-shot learning, or the capability of generating messages with prompting, and the fact that it does not require a resource-intensive fine-tuning process. As a result, we were able to start the message-generation process right away. While harnessing the message engine requires python coding, online resources with explanations of the coding made the process easy to learn (e.g., Theron, 2022; Tunstall et al., 2022; Wolf et al., 2020). The most time-consuming part was preparing for prompt engineering (e.g., scraping and cleaning the tweets, which took a few days to complete). However, now that the codebase exists, the prompt engineering process can be replicated quickly and made more flexible and comfortable for users. Once the final prompts were determined, the actual message generation process was fast. Loading the Bloom 7B1 model into Google Colab and generating 600 messages took a little over an hour. Overall, we found the model to be fast and efficient to use.

**Qualitative Characteristics of the Generated Messages and Computational Analyses (RQ2)**

Having affirmed the feasibility of the message engine, we will now discuss qualitative characteristics of the messages. First, the generated messages showed differing characteristics based on the structure of the prompts. When prompts included #folicacid, the generated messages also included hashtags and included features that are characteristic of tweet messages. For instance, the prompt *"Did you know, #Folicacid"* generated text such as *"Did you know, #Folicacid is one of the most important invitations in our daily diet?...So make sure to have it daily! #FolicAcid #VitaminB…"* On the other hand, when prompts did not include hashtags, the tone appeared to be more formal, and the content tended to include more factual information about folic acid. The prompt, *"Consuming folic acid is important because"*, for example, generated text such as *"Consuming folic acid is important because it helps prevent neural tube defects, which are birth defects in the brain and spinal cord. Folic acid is also necessary for healthy blood cells and nerve cells, and for maintaining healthy red blood cells…"*. These results point to how prompt technology can be used to strategically control qualities and tone of the AI-generated content, which is a fundamental feature.

Next, the results from the computational analysis showed that AI-generated and human-generated messages were similar in terms of word distributions, topics discussed, semantics, readability, and sentiment (see Table 1 and Figure 2). For instance, the uni- and bi-grams show that both AI-generated and human-generated messages contain the words related to neural tube and birth defects, focused generally on folic acid, pregnancy, and health topics. An analysis of text readability statistics (Flesch, 1953) showed that AI-generated messages were as easy to read, if not easier to read, compared to the human-generated messages ($m_{Flesch\text{-}Score\ AI}$ = 68.4, $m_{Flesch\text{-}Score\ human}$ = 63.4; $n.s.$). Next, a sentiment analysis using the valence-aware dictionary approach (VADER) showed no significant differences in terms of message sentiment, although nominally the AI-generated messages again ranked slightly higher (i.e., more positive sentiment; $m_{Vader\ Compound\ AI}$ = .25, $m_{FVader\ Compound\ human}$ = .23; $n.s.$). Overall, these analyses showed that there were no major differences between the AI-generated and human-generated messages.

|  | AI-Generated Messages | Human-Generated Messages | $t$ ($p$-value) |
| --- | --- | --- | --- |
| Flesch Reading Ease | 63.4 (17.9) | 68.4 (20.3) | .99 (.32) |
| Sentiment | .25 (.53) | .23 (.49) | .16 (.87) |

**Table 1:** Reading ease and sentiment analysis of AI-generated vs. human-generated messages. Means (standard deviations) and results of a t-test comparing the means of AI vs. human-generated messages for computational analyses of reading ease (Flesch Reading Ease, higher scores are more readable, scale ranges up to 1000) and message sentiment (Vader Compound Score, scores range from -4 to 4, with higher scores more positive).

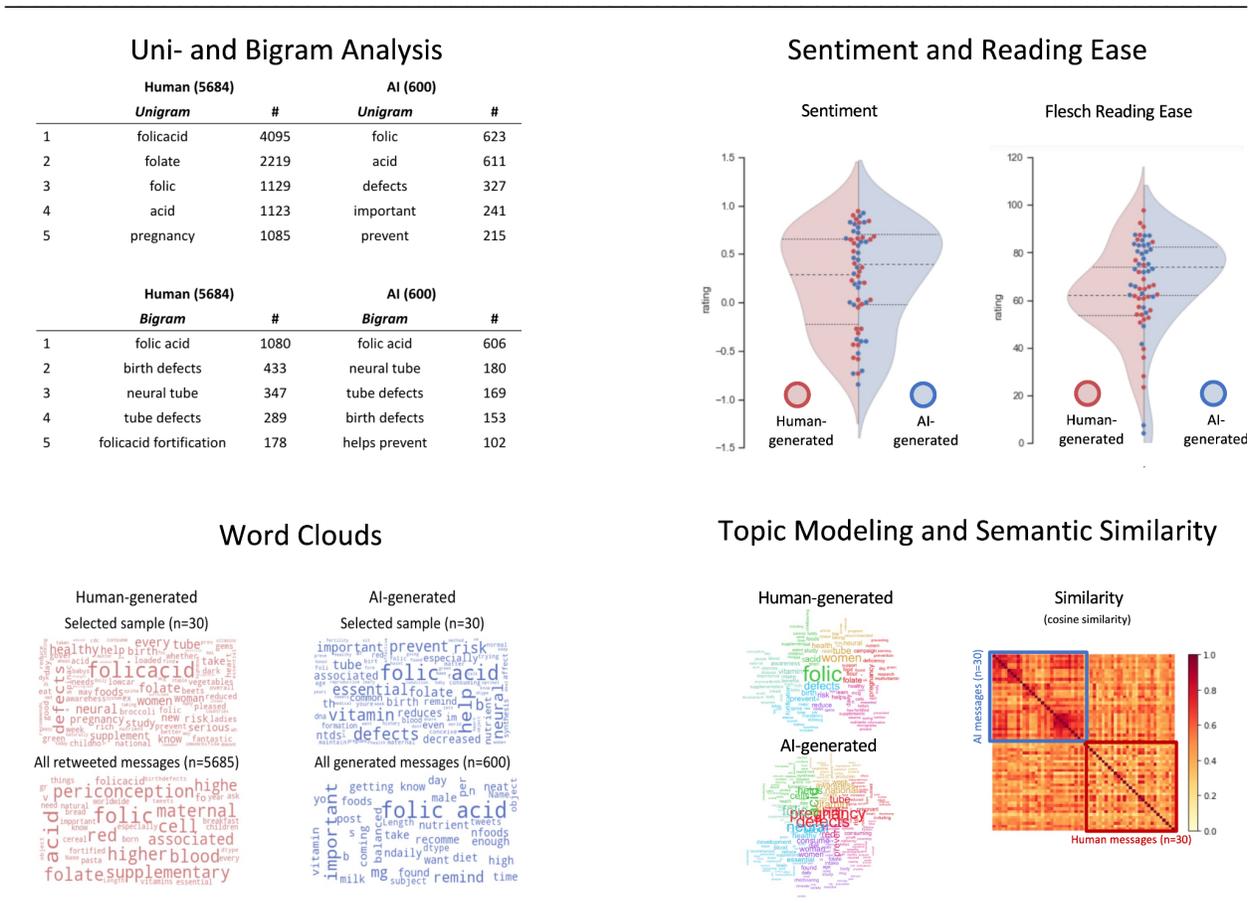

**Figure 2:** Results from computational analyses. Top left panel: NGram analysis. Top right panel: Results from sentiment and readability analysis, showing no significant differences between human- and AI-generated messages. Bottom left panel: WordClouds for selected and full samples of messages from human sources and the Bloom AI text generation system. Bottom right panel: Results from topic modeling (topic clouds) and semantic similarity analysis. The topic clouds illustrate the top words for each identified topic in a color-coded fashion. The semantic similarity analysis illustrates the cosine similarities between the per-sentence embedding vectors. The top 30 messages are from the AI generation, the bottom 30 messages represent the human generated ones.

Next, we focused on topics covered and semantic properties of the messages: A topic modeling analysis showed that both AI and human-generated messages discussed similar topics centered around pregnancy, birth defects, and essential vitamins. One observation is that the AI-generated messages appeared to weigh the topics of pregnancy and risk of birth defects more heavily. Lastly, we conducted an analysis of semantic similarities between all messages using the sentence-transformer package. This analysis asked how similar the sentence vectors of each of the 60 messages (30 AI-generated and 30 human messages) were to each other, as assessed via cosine similarity. As can be seen in Figure 2, the AI-generated messages were more similar to each other ($m_{similarity\ AI}$ = .59, $m_{similarity\ human}$ = .43; $p < 0.001$; note that while we removed prompts for other analyses, we included prompts here because the semantic similarity analysis requires entire sentences). As illustrated by Figure 2, a block-like structure separates the AI-generated messages from the rest, and within the AI messages, subblocks or similarity clusters exist. Importantly, all messages are fairly similar to each other in this semantics measure, with an average similarity score of about .5. In sum, these analyses confirm that the AI-generated messages were on topic, readable, and generally positive in tone.

**Results from the Human Evaluation Study (H1)**

The results from the one-sample *t*-test supported H1 (see Table 2). In fact, AI-generated messages were rated significantly higher than human-generated tweets for both dimensions - message clarity ($t_{clarity}$ = 4.32, $p$ < .01) and message quality ($t_{quality}$ = 5.39, $p$ < .01). Figure 3 also shows the result for each dimension for the full population as well as divided by gender. On average, the participants rated AI-generated and human-generated messages as relatively clear and easy to understand and containing quality content appropriate to increase FA awareness (ratings > 3 out of 5). However, the left graph in Figure 3 shows that the distribution of the AI-generated messages was concentrated toward the top (around 4) while human-generated messages had a slightly more even distribution, concentrated around 3-3.5.

|  | AI-Generated Messages | Human-Generated Messages | t (*p*-value) |
| --- | --- | --- | --- |
| Clarity | 3.77 (.55) | 3.22 (.43) | 4.32 (<.01) |
| Quality | 3.65 (.42) | 3.12 (.33) | 5.39 (<.01) |

**Table 2:** Results from the online survey that asked participants to rate AI-generated and human-generated messages in terms of clarity and quality. This shows the means and standard deviations (scales range from 1-5) and the results of the t-test comparing the ratings for AI vs. human-generated messages.

The distribution for females appeared wider than for males, with more AI-generated messages rated higher than human-generated messages. In addition, for males, AI-generated messages on average seemed similarly rated as the human-generated messages in terms of clarity. This may be the case because many FA messages referred to the importance of consuming the vitamin during pregnancy, which may not apply to college-age males. Finally, Figure 3 also combines clarity and quality ratings via a scatter plot, revealing that many AI-generated messages were rated high (around 4 and above) for both clarity and quality, thus making them the best candidate messages within the sample.

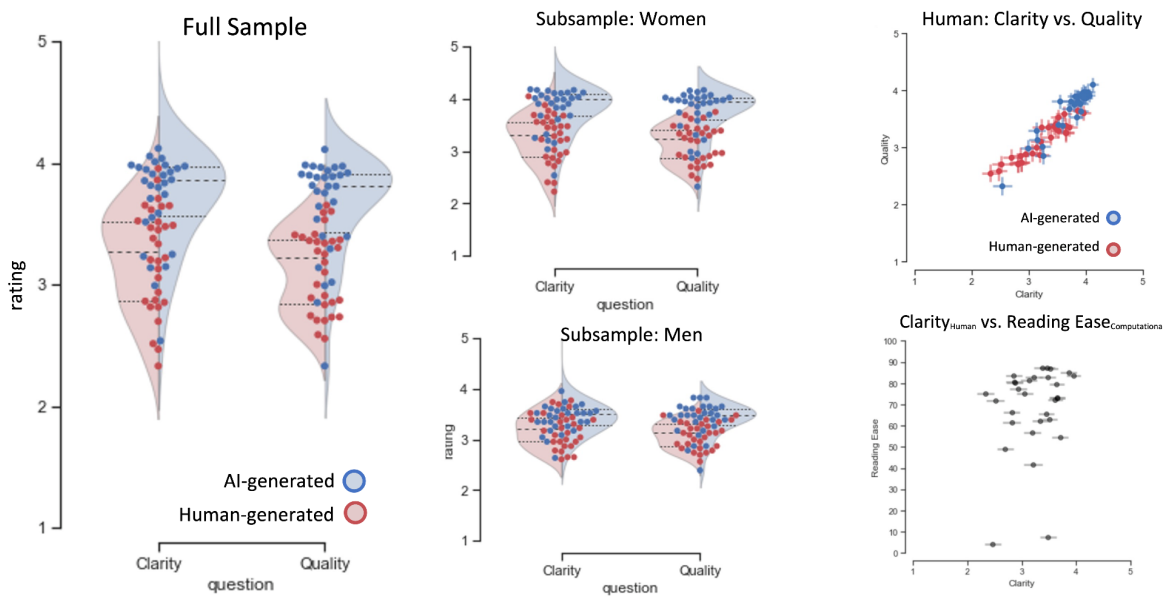

**Figure 3:** Additional results from human evaluation study. Left panel: The 30 AI-generated (blue dots) and 30 human-generated (red dots) were rated on a 5-point Likert scale for clarity and quality. Middle panels: Rating distributions divided by gender. Top right panel: Relationship between average human clarity and quality ratings for all individual messages. Bottom right panel: Relationship between human clarity ratings and computational reading ease scores by message (r = 0.23).

## Discussion

This study examined the feasibility of using the Bloom message engine to generate clear and high-quality awareness messages. The Bloom message engine is not only easy to harness but also has the potential to generate messages that humans perceive as easy to understand and appropriate to increase public knowledge about folic acid.

**Main Findings**

First, our study found that the Bloom message engine was easy to use. Even with moderate coding skills, the system can be utilized, and it would be feasible to create a turnkey solution for the message engine prototype. Second, the computational analyses demonstrated that the AI-generated messages contain high-quality information and that they are similar to the human-generated messages in specific quantitative text characteristics like word distributions, readability, sentiment, discussed topics, and general semantic similarity. One distinct feature was that AI-generated messages seemed to contain more words related to prevention. Perhaps this speaks to the differences in tweets vs. AI-generated messages. Messages may exhibit a tendency to have more positive sentiment and focus on the importance of folic acid (e.g., preventing neural tube defects). This all points to the enormous potential of the Bloom model in combination with prompt engineering, whose role in this field is likely to explode over the next few years.

Perhaps most importantly, humans rated the AI-generated messages as more clear and higher in quality compared to the retweeted human-generated messages. The fact the human-generated messages were retweeted means that those messages potentially contained elements that people thought were worthy to share with other people. The results showed that the Bloom message engine has the potential to generate informative awareness messages that surpass the retweeted messages. For obvious ethical reasons, we refrained from actually disseminating the AI-generated messages on social media, but we argue that it would be likely that they would also be shared more often. The potential to go viral is one of social media's most notable features when it comes to health communication because viral messages can reach billions of people across the globe to spread awareness about certain topics. In summary, the current results confirm that the Bloom message engine has a high potential to generate clear and high-quality content for awareness messages.

**Implications for Health Communication and Beyond**

The results of our study show that without additional training with topic-specific datasets (i.e., fine-tuning), the AI message engine can generate focused, clear, and appropriate awareness campaign messages. Since the system can generate hundreds of messages within a few hours, it can alleviate the burden of cost and manpower required in creating health communication messages. Moreover, the prompting technology allows the researchers to steer the content and sentiment of the generated messages (e.g., informal setting to fit social media, or formal to fit professional settings). This feature is extremely promising because it allows communicators to infuse strategy and theory into the AI generation process, which otherwise would remain a bit of a 'creative black box'. After the message generation step, health researchers could still filter through all messages to select and modify them or enter promising candidates back into the system as prompts to try to generate more of their kind. An important note here is that the message engine is not meant to be the decision-maker. Rather, it is a tool that is meant to support the health researcher to save cost, manpower, and time, and to add a replicable, but creative element to the generation process. In all practical use cases, ethics and legal aspects would obviously still require the presence of humans as curators of content.

Another application area of the AI-message engine is that it can be used to generate messages for topics beyond simple awareness messaging (Baclic et al., 2020). As argued above, we deliberately focused on awareness as a fundamental goal of health communication (Bettinghaus, 1986; McGuire et al., 2001), but clearly, it will be promising to explore the potential of this approach to influence health-related attitudes and behaviors. Furthermore, given the rise of image generation as a visual complement to text generation, it would also be interesting to start generating health-related imagery and combine such messages with AI-generated texts (Crowson et al., 2022). This is especially important because newer social media platforms are increasingly depending on visuals (e.g., Snapchat, TikTok). Overall, we anticipate that this work will find many applications in health communication in the near future, and similar cases could be made for other communication topics (e.g., environmental and political communication) or applied business communication (e.g., AI-copywriting systems, which are already offered commercially).

**Potential of AI-Message-Generation to Contribute to Communication Science and Theory**

Going beyond the obvious application potential, we next ask about the theoretical potential and contributions of this work. At first glance, it might be tempting to view AI message generation as a methodological innovation only, but we argue that the advent of natural text generation systems holds immense potential for method-theory synergy (Greenwald, 2012; Weber et al., 2018): Specifically, a healthy mix of *theory*, *measurement*, and *control* is central to the progress of any science (Bechtel, 2008; Chalmers, 2013; Craver & Darden, 2013). Put simply, our theories affect the phenomenon we can conceive of and use to explain the phenomenon in question. Our measures, in turn, determine what phenomena we can 'see' and with which precision. Lastly, the ability to control lets us 'push things around', manipulating or intervening causally on variables to test our theories. In physics, chemistry, and biology, the triad of *theory, measurement,* and *control* enabled everything from understanding atoms and genetics (on the basic science side) to building mechanical engines and synthetically assembled vaccines (on the applied side). In NLP, the advent of powerful theories that enable machines to generate natural text clearly represents a breakthrough in the quest to understand, measure, and control linguistic or even cognitive capacities (LeCun et al., 2015; Mitchell, 2019). As we saw from the results of this study, this advancement from powerful theories in NLP has many implications for communication (Dubova, 2022; Hassabis et al., 2017; Lake et al., 2016).

Critically, however, we are not claiming that these advances have solved long-standing theoretical questions about the nature of language, let alone symbolic communication - it's quite the opposite. So far, progress in language modeling has focused on language-intrinsic aspects (e.g. syntax and semantics; Bender & Koller, 2020), whereas pragmatic and extralinguistic processes that are critical to communication have been almost ignored. However, the verbal distinction between semantics and pragmatics has always been cumbersome since the underlying phenomena appear to gradually blend into each other. The current work thus represents a step to bridge between NLP and communication, albeit only an initial one. Moreover, the fact that human participants evaluated the AI-generated messages as even better than human-generated ones (and as better as even the most highly shared human-generated messages) underscores the importance of this approach.

In addition to potentially surpassing human-level performance for message generation, another theoretical benefit of AI systems is that they are rigorously quantitative. While it is often claimed that deep learning is like a 'black box' (Chen et al., 2020; Marcus & Davis, 2019; Xu et al., 2019), this is actually not the case: Rather, each of Bloom's over seven billion parameters are computed and thus, in principle, objectively knowable[6]. This again points to a key feature of science because it affords better measurement, better control, and ultimately better theoretical understanding. Indeed, one could perhaps compare AI models to the 'Petri dish' in chemistry, i.e., as providing a precisely controlled experimental setting in which researchers can generate synthetic messages under controlled laboratory conditions, and later test how they fare in the market. This promotes insights into properties that make messages effective, an issue that still remains insufficiently understood (Harrington, 2015, 2016; O'Keefe & Hoeken, 2021). In this sense, especially the prompt engineering process, which has been successfully applied here, has barely scratched the surface of what is possible - and needed - to fully realize the potential of this applied fruit of a theoretically matured science of messages.

---

[6] Granted, this mindboggling complexity prevents us from understanding the system behavior intuitively, which is what people mean when they call it a black box. However, except for deliberately random parameters, the systems behave deterministically, and, in this sense, the black box metaphor is mistaken.

**Ethical Considerations and Limitations**

The emergence of AI-based methods for communication content generation clearly raises important ethical issues. Examples of existing issues include bots in the political domain (which spread misinformation), online troll farms (which stoke conflict), or general misuse of persuasive technologies for target advertising or mood, and emotion intervention (Kreps et al., 2020; Solaiman et al., 2019). At this point, there is no clear regulatory framework for these technologies. Although a few 'responsible AI' initiatives exist, the novelty of the topic and the evolving nature of the underlying science prevent any final judgment. However, the creators of Bloom have attempted to address this issue by releasing Bloom under a RAIL license (responsible AI license, Contractor et al., 2020), which includes a number of disallowed use cases, like generating law-violating content. Among these cases is using Bloom for medical purposes. Clearly, given the language model's lack of medical knowledge (Schmälzle & Wilcox, 2022), it would be irresponsible to use the model to generate diagnoses and prescribe treatment. On the other hand, using the model to improve general health awareness information seems like a particularly beneficial use case for how language models (under supervision by health communication experts acting as message curators) could be used for social good and to improve the quality of public health information. With this in mind, we also want to underscore the importance that communication researchers engage with and help shape this debate.

Like all research, the current paper has several limitations. First, we tested only one topic, folic acid. Although this was an informed choice based on the characteristics of this particular health topic, it will be necessary to expand these findings to other domains. Second, we focused only on health messages designed to raise awareness but ignored more downstream topics like attitude and behavior change. Again, we argue that this provides a logical starting point, but that more work needs to be done to expand to other topics within health communication. Relatedly, we only generated relatively short messages, such as the ones we find on social media (primarily twitter). However, health communication takes many forms, and as such more work is needed to generate e.g., health-related stories (longer text) or images (no text at all, but a different modality). Third, the computational and human evaluations of the generated messages could be improved. Regarding the computational analyses, we only focused on a number of salient and quantifiable characteristics, like NGrams, sentiment, and reading ease. However, there are additional topics that were not explored, such as similarity of the entire corpora of human and AI-generated messages, other theoretical concepts (e.g. use of politeness instead of general sentiment (Yeomans et al., 2019), or the use of specific kinds of persuasive strategies (Armstrong, 2010; O'Keefe & Hoeken, 2021; Tan et al., 2016). Again, this limitation is explained by the early stage of this research - this is only the third study of this kind that we are aware of - and we look forward to future work expanding the scope of computational message analytics. Finally, human evaluation is not without limitations either. In particular, one could criticize the narrow measures and the sample. Regarding the latter, we argue that the population we sampled from (college students) overlaps with the target population for a FA awareness campaign (potential parents and particularly women of childbearing age), but it would be advisable to also test how broader audiences respond to these messages.

## Summary and Conclusion

At the start of this paper, we asked readers to guess which of the two messages came from an AI. The answer is that it was the first message, and the fact that most readers could only guess the correct answer affirms our main findings. Specifically, the Bloom message engine was easy to use, the generated messages were generally on par with human messages in terms of quantitative characteristics, and they were rated as clearer and of higher quality compared to even the most retweeted human messages. Thus, the message engine could be used to alleviate the bottleneck in the message generation process for health awareness messages. Overall, this approach offers fruitful ground to use quantitative methods to examine the generation, content, and reception of messages.